%% file: Main.tex
\documentclass[10pt,twocolumn,letterpaper]{article}

\usepackage{iccv}
\usepackage{times}
\usepackage{epsfig}
\usepackage{graphicx}
\usepackage{amsmath}
\usepackage{amssymb}
\usepackage{float}
\usepackage{multirow}
\usepackage{bbm}
\usepackage{everyshi}
\input{math_commands.tex}
\usepackage[pagebackref=true,breaklinks=true,colorlinks,bookmarks=false]{hyperref}

\iccvfinalcopy % *** Uncomment this line for the final submission

\def\iccvPaperID{3} % *** Enter the ICCV Paper ID here

\ificcvfinal\fi

\begin{document}

%%%%%%%%% TITLE
\title{Kinship Representation Learning with Face Componential Relation}

% \author[1]{Wen-Tai Su}
% \author[2]{Min-Hung Chen}
% \author[2]{Chien-Yi Wang}
% \author[3]{Shang-Hong Lai}
% \author[4]{Trista Chen}
% \affil[1]{Novatek}  \affil[2]{NVIDIA}
% \affil[3]{National Tsing Hua University}
% \affil[4]{Microsoft Corporation}

% \author{Wen-Tai Su\inst{1}\index{Su, Wen-Tai} \and
% Min-Hung Chen\inst{2} \and
% Shang-Hong Lai \inst{1}\and 
% Trista Chen\inst{1}}

% \institute{Dept. EE, National Tsing Hua University, Hsinchu 300044, Taiwan \\
% \and
% Dept. ECE, University of California, Los Angeles, USA\\
% }
\author{Weng-Tai Su\thanks{Work was done during Microsoft}\\
Novatek Microelectronics Corp.\\
% (\small Work was done during Microsoft)\\
% {\tt\small wengtai2008@hotmail.com}
% For a paper whose authors are all at the same institution,
% omit the following lines up until the closing ``}''.
% Additional authors and addresses can be added with ``\and'',
% just like the second author.
% To save space, use either the email address or home page, not both
\and
Min-Hung Chen\footnotemark[1]\\
NVIDIA\\
% (\small Work was done during Microsoft)\\
% {\tt\small vitec6@gmail.com}
\and
Chien-Yi Wang\footnotemark[1]\\
NVIDIA\\
% (\small Work was done during Microsoft)\\
% {\tt\small chienyiw@nvidia.com}
\and
Shang-Hong Lai\footnotemark[1]\\
National Tsing Hua University\\
% {\tt\small lai@cs.nthu.edu.tw}
\and
Trista Chen\\
Microsoft Corp.\\
% {\tt\small trista.chen@gmail.com}
}
\newcommand{\fix}{\marginpar{FIX}}
\newcommand{\new}{\marginpar{NEW}}

\newcommand{\MainMethodAbbr}{FaCoRNet}
\newcommand{\MainMethod}{Face Componential Relation Network}
\newcommand{\MainFusion}{Face Componential Relation}
\newcommand{\MainFusionAbbr}{FaCoR}
\newcommand{\CA}{Channel Interaction}
\newcommand{\ATT}{Relation Guidance}
\newcommand{\Learning}{Relation-Guided Contrastive Learning}
\newcommand{\LearningLoss}{Relation-Guided Contrastive Loss}
\newcommand{\LearningAbbr}{Rel-Guide}

\def\I{{\mathbf I}}
\def\X{{\mathbf X}}

\maketitle
% Remove page # from the first page of camera-ready.
\ificcvfinal\thispagestyle{empty}\fi

%%%%%%%%% ABSTRACT
\begin{abstract}
Kinship recognition aims to determine whether the subjects in two facial images are kin or non-kin, which is an emerging and challenging problem. However, most previous methods focus on heuristic designs without considering the spatial correlation between face images.
% Traditional methods only focus on learning discriminative features of each facial image from the paired samples, while neglecting how to fuse the two obtained facial image features and components the relationship between them. 
In this paper, we aim to learn discriminative kinship representations embedded with the relation information between face components. 
To achieve this goal, we propose the \textit{\textbf{Fa}ce \textbf{Co}mponential \textbf{R}elation \textbf{Net}work (\textbf{\MainMethodAbbr})}, which learns the relationship between face components among images with a cross-attention mechanism, to automatically learn the important facial regions for kinship recognition. Moreover, we propose \Learning, which adapts the loss function by the guidance from cross-attention to learn more discriminative feature representations. The proposed \MainMethodAbbr~outperforms previous state-of-the-art methods by large margins for experiments on multiple public kinship recognition benchmarks. Our code is available at \href{https://github.com/wtnthu/FaCoR}{\textcolor{blue}{https://github.com/wtnthu/FaCoR}}.
% Our code is available at \href{https://github.com/wtnthu/FaCoR}{\textcolor{blue}{https://github.com/wtnthu/FaCoR}}.
% The code will be released upon acceptance.

\end{abstract}

%%%%%%%%% BODY TEXT
\section{Introduction} 
\label{sec:introduction}
\input{1_introduction}

\section{Related Work} 
\label{sec:related}
\input{2_related}

% \section{Method: Attention-Guided Facial Structure Learning} 
\section{Proposed Methods} 
\label{sec:mehhods}

\input{3_proposed}

\section{Experiments} 
\label{sec:experiment}
\input{4_experiment}

\section{Conclusion and Future Work} 
\label{sec:conclusion}
\input{5_conclusion}

{\small
\bibliographystyle{ieee_fullname}
\bibliography{ref}
}

\end{document}

%% file: math_commands.tex
\usepackage{amsmath,amsfonts,bm}

\def\eqref#1{equation~\ref{#1}}
\def\1{\bm{1}}

\def\rr{{\textnormal{r}}}

\DeclareMathAlphabet{\mathsfit}{\encodingdefault}{\sfdefault}{m}{sl}
\SetMathAlphabet{\mathsfit}{bold}{\encodingdefault}{\sfdefault}{bx}{n}

\newcommand{\Conv}{\mathrm{Conv}}
\def\I{{\mathbf I}}
\def\X{{\mathbf X}}
\def\F{{\mathbf F}}

\def\rr{{\mathbf r}}

%% file: 1_introduction.tex
\begin{figure}[htbp]
\centering
\includegraphics[width=0.45\textwidth]{
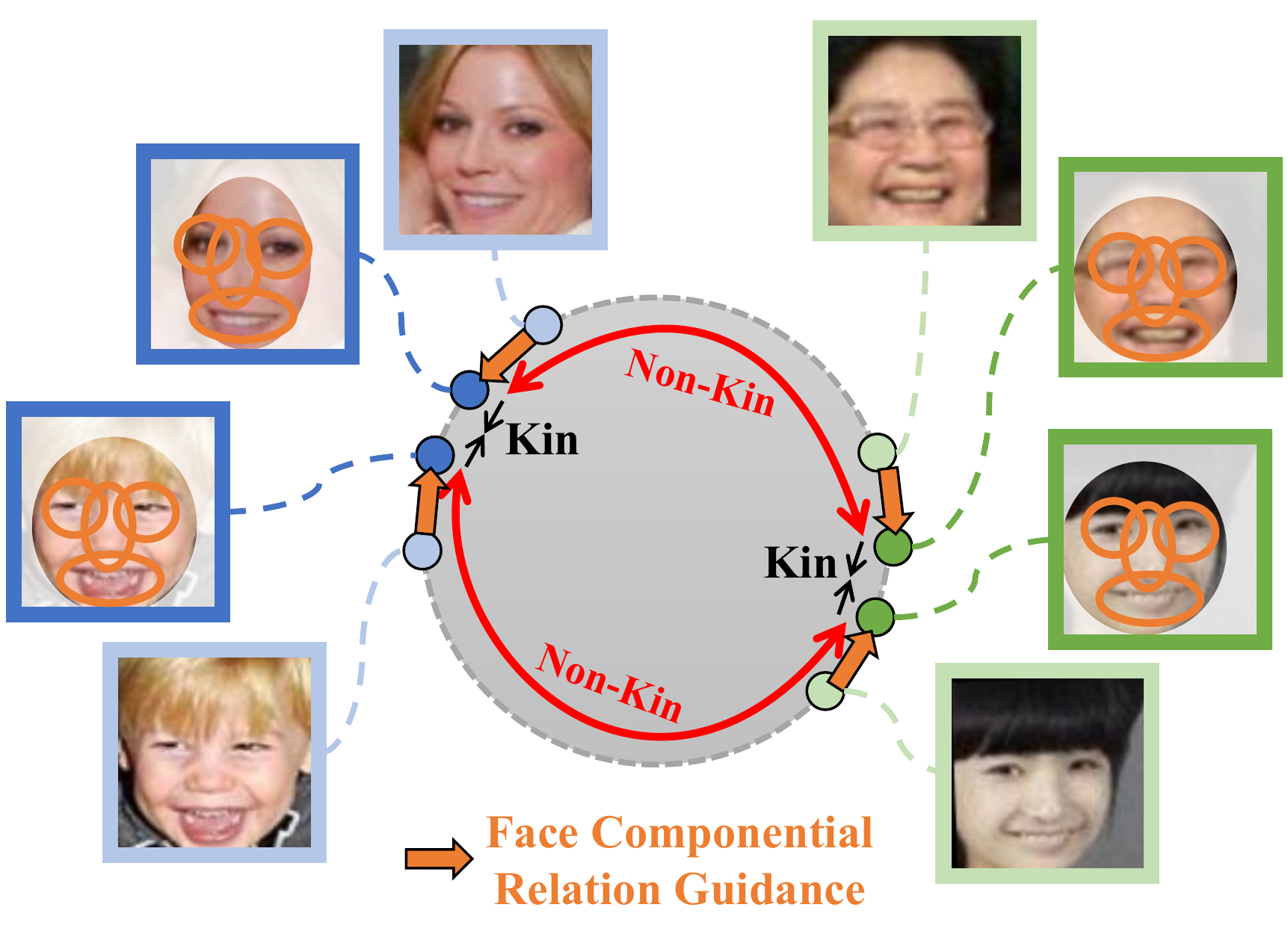}
\caption{
Our method uses face components as clues and guides the training with the relation of facial image pairs, where the relation estimation for face components (darker blue and green colors) can further pull \textit{kin} faces together and push away \textit{non-kin} faces, improving the efficacy of contrastive learning from the original whole-face features (lighter blue and green colors).
} 
\label{fig:teaser}	
\end{figure}

In recent years, \textit{kinship recognition}, which aims to determine whether a given pair of face images have a kinship relation, has attracted public attention. Kinship recognition is inspired by the biological discovery \cite{dal2010lateralization} that the appearance of a human face implies clues about kinship-related information. It can be widely used in various scenarios including missing child search \cite{lu2013neighborhood}, automatic album organization \cite{zhou2012gabor}, child adoption \cite{yan2014discriminative}, and social media applications \cite{dehghan2014look}. 
Facial kinship recognition includes both face representation learning and face similarity matching, where the former aims to learn discriminative features for input facial images, and the latter is to design models to predict the kin/non-kin relationship between images in a pair. 
The main challenges of kinship are mixed variations due to an uncontrolled environment, such as the large gap in age, expression, pose, illumination, etc. Under these variations, it is challenging to learn representations that can help discover genetic relationships between two samples from facial appearance and identify hidden similarities inherited from genetic connections between different identities.

To deal with these challenges, several traditional approaches incorporate hand-crafted features \cite{lu2013neighborhood} with metric learning \cite{fan2020efficient} to learn discriminative features.
Motivated by the success of deep learning, various methods improve kinship recognition by exploiting powerful deep feature representations.  CNN-Point~\cite{zhang122015kinship} first adopts a CNN model to extract discriminative features, outperforming previous hand-crafted ones. For the extension, several CNN-based approaches \cite{dahan2020unified, luo2020challenge, yu2021deep} focus on designing fusion mechanisms to integrate the features among an image pair. 
Recently, the supervised contrastive approach \cite{zhang2021supervised} learns discriminative features by contrastive loss, which achieves state-of-the-art performance in kinship recognition.
However, the existing approaches have several issues. First, most methods directly exploit feature vector representations, ignoring spatial correlation within face images. Moreover, most of the approaches rely on heuristic designs. For example, the feature fusion approaches \cite{yu2020deep, zhang122015kinship} utilize several arithmetic combinations or feature concatenation to fuse the feature pair for kinship recognition. Despite the state-of-the-art performance from \cite{zhang2021supervised}, the results are sensitive in the hyperparameter setting of the contrastive loss.

To address the above issues, let us first think again: \textit{How do humans recognize kinship relationships?} To recognize accurately, humans usually first compare several biological \textbf{\textit{face components}} of two people, such as eye color, nose size, cheekbone shape, etc., and then analyze the \textbf{\textit{relation}} between these comparisons. 
For example, if the noses in the image pair appear similarly, then there is a higher chance that this is a \textit{kin} pair.
Therefore, we adopt this idea, focusing on how to exploit these \textbf{\textit{face components}} to learn the \textbf{\textit{relation}} between images in a pair, where clues from \textit{face components} can infer the genetic relationships between them. 
In this work, we aim to learn discriminative feature representations embedded with face component information, without a strong reliance on heuristic designs, as shown in Fig.~\ref{fig:teaser}.

To achieve the abovementioned goal, we first propose the \textit{\MainFusion~(\MainFusionAbbr)} module to learn the relation between images in a pair with the consideration of face components. The feature representations are then enhanced with the cross-relation between face components (e.g., eyes, nose, mouth, etc.) which are critical to kinship recognition.
Moreover, we propose the novel \textit{\LearningLoss~(\LearningAbbr)} based on cross-attention estimation instead of heuristic tuning \cite{zhang2021supervised}. The attention map can control the degree of penalty in the loss function, which can let the feature representation of kin relation get closer in the feature space. In other words, it penalizes the hard samples to learn more discriminative features for kinship recognition.  
The whole architecture is named \textit{\textbf{Fa}ce \textbf{Co}mponential \textbf{R}elation \textbf{Net}work (\textbf{\MainMethodAbbr})}.

The experimental results show that our \MainMethodAbbr~achieves SOTA performance on the largest public kinship recognition benchmark, FIW \cite{robinson2021survey}. To be specific, our work outperforms the previous best method in three tasks with standard protocol by 2.7$\%$ (79.3$\%$ $\rightarrow$ 82.0$\%$) in the kinship verification task, 0.7$\%$ (84.4$\%$ $\rightarrow$ 85.1$\%$) in the tri-subject verification task, and 14.2$\%$ (40.0$\%$ $\rightarrow$ 54.2$\%$) in the search and retrieval task. We also show that our \MainMethodAbbr~achieves SOTA performance on the other two widely-used kinship recognition benchmarks, KinFaceW-I and KinFaceW-II.

Our contributions are summarized as follows:

\begin{itemize}
\item We propose a novel \textit{\textbf{Fa}ce \textbf{Co}mponential \textbf{R}elation \textbf{Net}work (\textbf{\MainMethodAbbr})}

that learns relevance from the face components of image pairs with the cross-attention mechanism, and adaptively learns important face components for kinship recognition.

\item We propose a novel \textit{\LearningLoss} that embeds cross-relation estimates to guide the contrastive loss without heuristic tuning, which controls how hard samples are penalized during training.

\item The proposed \textbf{\textit{\MainMethodAbbr}} model outperforms previous SOTA methods by large margins on multiple standard kinship recognition benchmarks.

\end{itemize}

%% file: 2_related.tex
In the past few years, several kinship recognition approaches have been proposed \cite{chen2022deep, dahan2020unified, fan2020efficient, hormann2020multi, huang2022adaptively, li2021meta, lin2021styledna, lu2013neighborhood, luo2020challenge, serraoui2022knowledge, shadrikov2020achieving, song2020kinmix, yu2021deep, zhang122015kinship, zhang2021supervised}, where most of them focus on extracting discriminative feature for each facial image. Traditional approaches include designing hand-crafted feature extractors~\cite{abdi2010principal, cui2013fusing, somanath2012can}
and metric learning \cite{dibeklioglu2017visual, ding2017trunk, huang2022adaptively} for solving similarity metrics in kinship recognition. 
Recently, deep learning methods make significant advances, including
two main categories: \textit{feature fusion} and \textit{deep metric learning}.

\noindent\textbf{Feature Fusion:} 
\cite{zhang122015kinship} utilizes the multiple face regions as the model inputs to learn richer facial features for kinship recognition. The multi-task deep learning-based approach \cite{dahan2020unified} uses seven kinship sub-classes to jointly train with the kinship labels for kin recognition. Ustc-nelslip \cite{yu2020deep} adopts a siamese network to extract features and designs three different math operations to fuse feature pairs, followed by direct concatenation with a fully-connected layer. \cite{serraoui2022knowledge} proposes an advanced knowledge-based tensor similarity extraction framework for automatic facial kinship verification that utilizes four pre-trained networks to improve the performance.

\noindent\textbf{Deep Metric Learning:}  \cite{duan2017face} proposes coarse-to-fine transfer to capture kinship-specific features from faces using supervised coarse pre-training and domain-specific retraining paradigms. The contrastive learning approach \cite{zhang2021supervised} utilizes supervised contrastive loss with the ArcFace pre-trained model \cite{deng2019arcface} and two MLP layers to learn more robust features in the training stage. For the evaluation, it removes the MLP layers and extracts the middle-layer backbone features to evaluate the cosine similarity to determine the kinship relation in an image pair, thus achieving state-of-the-art performance for kinship recognition.  \cite{huang2022adaptively} presents a novel cross-pair metric learning approach that introduces a k-tuplet loss. This approach effectively captures both low-order and high-order discriminative features from multiple negative pairs.

The main issues of the above methods are that most methods rely on heuristic designs, and directly exploit feature vector representations, ignoring spatial correlation within face images. 
Different from the above approaches, our proposed \MainMethodAbbr~considers how to use face components to learn the correlation between image pairs, and find out important facial parts for kinship recognition. Moreover, our approach incorporates the face componential correlation
to adapt contrastive learning automatically, without a strong reliance on heuristic designs.

\begin{figure*}[!ht]
\centering
\includegraphics[width=0.95\textwidth]{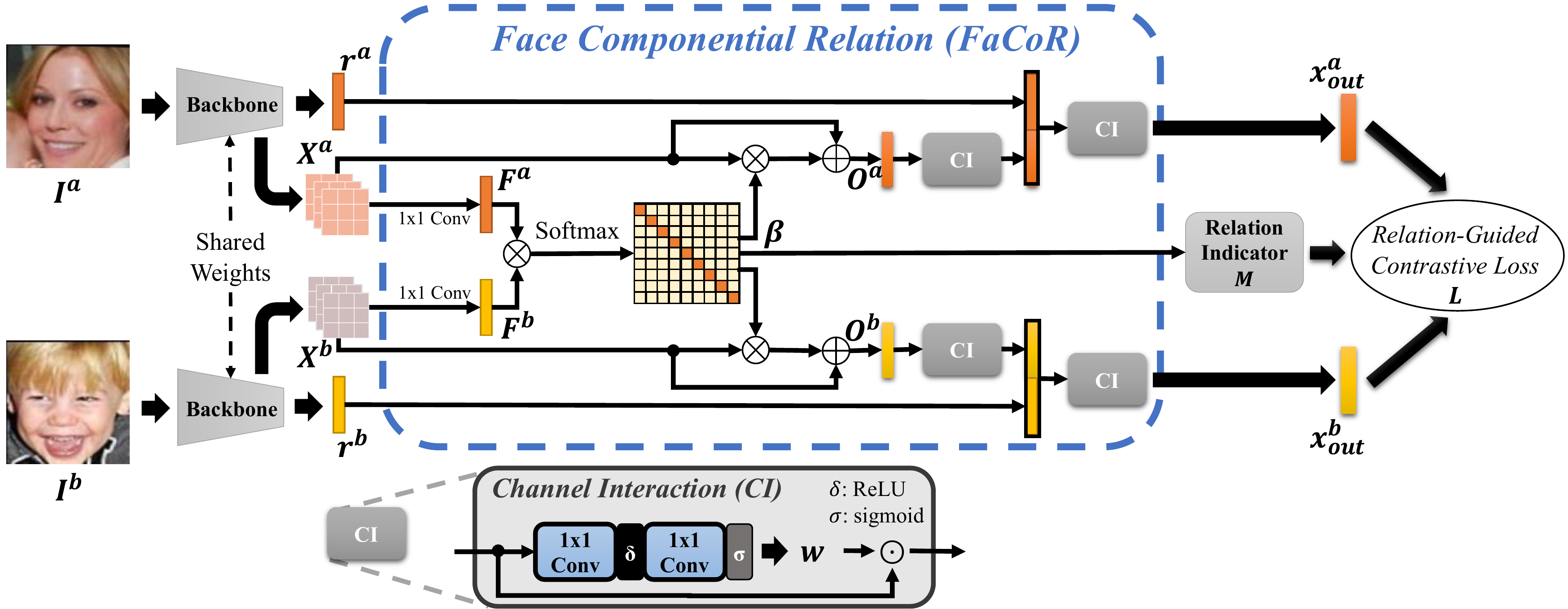}
]\caption{An overview of the proposed \textit{\MainMethod~(\MainMethodAbbr)} consisting of a backbone and the \MainFusion~(\MainFusionAbbr) module (Sec.~\ref{sec:MainFusion}),
trained with the \LearningLoss~$L$ (Sec.~\ref{sec:Learning}).
In the \MainFusionAbbr~module, we compute the cross-attention features $\mathbf{O}$ embedded with face componential relation and then fuse the information from high-level features $\rr$ via \CA~(CI) blocks. During training, the attention map $\boldsymbol{\beta}$ is adopted as the guidance to learn discriminative representations.
} 
\label{fig:network}	
\end{figure*}

%% file: 3_proposed.tex
In this work, we propose the \textit{\textbf{Fa}ce \textbf{Co}mponential \textbf{R}elation \textbf{Net}work (\textbf{\MainMethodAbbr})}, which considers the \textit{face components} and learns the cross-\textit{relation} between face images in a pair to benefit kinship recognition.

\MainMethodAbbr~consists of a shared-weights backbone 
that extracts features as the inputs to the \textit{\MainFusion~(\MainFusionAbbr)} module.
\MainFusionAbbr~is an attention-based module that computes the cross-relation among a face image pair and enhances feature representations to fully exploit the symmetry of face components in the image pair. In addition, cross-layer features are mutually interacted and fused in the channel dimension by the \CA~(CI) blocks  (Sec.~\ref{sec:MainFusion}).

Moreover, the proposed \textit{\LearningLoss} utilizes the computed cross-relation to guide the contrastive loss, facilitating learning of more discriminative representations for kinship recognition (Sec.~\ref{sec:Learning}).

The overall framework is illustrated in Fig.~\ref{fig:network}.

\subsection{\MainFusion} \label{sec:MainFusion}
One core question for kinship recognition is: \textit{How to properly extract and compute the relation between face components in a face image pair?} However, most existing methods are not designed for the face components of kinship recognition. To solve this, we propose the \textit{\MainFusion~(\MainFusionAbbr)} module, which can embed the relation information between face components into kinship feature representations, as the core component of our \MainMethodAbbr~(Fig.~\ref{fig:network}).

We denote the input face image pair as $(\I^{a}, \I^{b}) \in \mathbb{R}^{h \times w \times 3}$, the extracted feature maps from the backbone's middle-layer as $(\X^{a}, \X^{b}) \in \mathbb{R}^{H \times W \times C}$, and the high-level features from the backbone's final layer as $(\rr^{a}, \rr^{b}) \in \mathbb{R}^{C}$, where $H$, $W$, and $C$ represent the height, width, and the channel number of feature maps, respectively. 
The proposed \MainFusionAbbr~module mainly serves two purposes: 1) To adaptively learn the correlation between face image pairs, and 2) to learn the dependencies in face components between image pairs. These two directions help to learn which facial parts are important for kinship recognition.
More specifically, We first extract features $(\X^{a}, \X^{b})$ from the shared-weights backbone and then

use $1 \times 1$ convolution $\Conv$ to extract two intermediate flattened feature vectors $(\F^a, \F^b) = (\Conv_{1\times1}(\X^{a}), \Conv_{1\times1}(\X^{b}))$ $\in \mathbb{R}^{H \times W \times C}$.
Then, we find wide-range dependencies between the flattened feature vector pair $(\F^a, \F^b)$ and estimate the cross-attention map $\boldsymbol{\beta}$ as:
\begin{equation}
    \beta_{j,i} = \frac{\exp(s_{ij})}{\sum_{i=1}^{N}\exp(s_{ij})}~, ~s_{ij}= (\F_i^{a})^{T} \F_j^b ,
\label{eq:correlation}
\end{equation}
where $\beta_{j,i}$ estimates model attention in the $i$-th location of the $j$-th region.

We then multiply each output of the attention map $\boldsymbol{\beta}$ with the feature map $(\X^{a}, \X^{b})$ and adopt a learnable $\gamma$-scaled residual connection
to obtain the cross-attention features $(\mathbf{O}^{a}, \mathbf{O}^{b}) \in \mathbb{R}^{C \times HW}$, given by:

\begin{equation}
    \left( \mathbf{O}^{a}_j, \ \mathbf{O}^{b}_j \right) = \left(\X^{a} + \gamma \sum_{i=1}^{N} \beta_{j,i} \X_i^{a}, \  \X^{b} + \gamma \sum_{i=1}^{N} \beta_{j,i} \X_i^{b}\right).
\label{eq:self}
\end{equation}

All the operations are differentiable since they are purely linear and properly reshaped.

To effectively fuse the information from the cross-layer features, including high-level features $(\rr^{a}, \rr^{b})$ and the cross-attention features $\left(\mathbf{O^a}, \mathbf{O^b}\right)$,
we adopt \CA~(CI) blocks that encode inter-channel relations as shown in the gray block in Fig.~\ref{fig:network}. CI computes the interaction weights $\mathbf{w}$ via two sets of $1\times 1$ convolution, a sigmoid, and a ReLU activation function as follows:
\begin{align}
    \mathbf{w} = \sigma \left( \mathrm{Conv}_{1\times1}\left( \delta \left( \mathrm{Conv}_{1\times1}(\hat{\mathbf{x}}) \right) \right) \right),
\label{eq:channel_comp}
\end{align}
where $\mathrm{Conv}_{1\times1}(\cdot)$ is a $1\times 1$  convolution operation, $\sigma$ is the sigmoid operation, and $\delta$ is the ReLU operation. $\hat{\mathbf{x}}$ denotes the input to the CI block, where the elements in $\hat{\mathbf{x}}$ are multiplied element-wise with their corresponding weights to produce a set of weighted feature values $\mathbf{w} \hat{\mathbf{x}}$. 
Finally, the outputs of the \MainFusionAbbr~ module $(\mathbf{x^a_{out}}, \mathbf{x^b_{out}})$ are generated by fusing the information of cross-layer features via the \CA~blocks as follows:
\begin{equation}
    \left( \mathbf{x^a_{out}}, \mathbf{x^b_{out}} \right) = \left( \mathbf{CI} \left(\mathbf{\mathbf{CI} \left( \mathbf{O}^{a}\right)} \ || \ \rr^{a} \right),  \mathbf{CI} \left(\mathbf{\mathbf{CI} \left( \mathbf{O}^{b}\right)} \ || \ \rr^{b} \right) \right),
\label{eq:final}
\end{equation}
where
the operation $||$ denotes the concatenation of two feature maps in the channel dimension.

\subsection{\Learning}\label{sec:Learning}
Contrastive learning \cite{chen2020simple, khosla2020supervised} is known as an effective representation learning approach. It allows the model to learn the discriminative features from data similarities and dissimilarities, even without labels. The supervised contrastive \cite{zhang2021supervised} approach learns more robust features in kinship recognition, achieving state-of-the-art performance. The main idea of contrastive learning is to learn the discriminative feature, where feature representations of kin relations in feature space would be close. Otherwise, the feature representations of non-kin relations in feature space are far apart. For the standard contrastive learning, given $N$ positive samples ($x_i, y_i$), the contrastive loss $L$ is given by:
\begin{align}
    L = \frac{1}{2N}  \sum_{i=1}^{N} \left( L_c(x_i, y_i) + L_c(y_i, x_i) \right)
\label{eq:contrastive1}
\end{align}
and $L_c(x_i, y_i)$ is defined as: 

\begin{align}
     L_c(x_i, y_i) = -log \frac{e^{sim(x_i, y_i)/\tau}}{\sum_{j=1}^{N} \mathbbm{1}_{[j \neq i] } (e^{sim(x_i, x_j)/\tau} + e^{sim(x_i, y_j)/\tau})},
\label{eq:contrastive2}
\end{align}

\noindent where $\mathbbm{1}_{[j \neq i]} \in \{0,1\}$ represents an indicator function that evaluates to 1 iff $j \neq i$, the negative samples are generated by incorporating positive from different kinship categories (i.e., $(x_i, x_j)$), and $sim(x, y)$ is the cosine similarity operation between $x$ and $y$. 

However, the kinship recognition performance of contrastive learning is sensitive to hyper-parameter $\tau$ \cite{zhang2021supervised}, which controls the degree of penalty for hard samples. 
To solve this problem, we propose the \textit{\LearningLoss~(\LearningAbbr)} with a relation indicator $\mathbf{M}$, which guides the contrastive loss with the cross-attention estimation instead of heuristic tuning,
as shown in Fig.~\ref{fig:network}.

The main idea is that a smaller value from the cross-attention map needs a greater degree of penalty for hard samples. In other words, the small correlation between image pairs in kin relation 
needs a greater degree of penalty. This idea is also similar to updating the network with a large gradient to improve kinship recognition performance, and vice versa. Therefore, we extract the cross-attention map $\boldsymbol{\beta}$ in Eq.~\ref{eq:correlation}, which corresponds to the face component correlation between image pairs. Then, we utilize the relation indication function $\mathbf{M}$ to estimate the similarity value ${\psi}$ to replace the fixed value $\tau$ in Eq.~\ref{eq:contrastive_m} as: 

\begin{align}
     L_c(x_i, y_i) = -log \frac{e^{sim(x_i, y_i)/{\psi}}}{\sum_{j=1}^{N}\mathbbm{1}_{[j \neq i] }(e^{sim(x_i, x_j)/{\psi}} + e^{sim(x_i, y_j)/{\psi}})},
\label{eq:contrastive_m}
\end{align}

\noindent where ${\psi}=\mathbf{M}(\boldsymbol{\beta})/s$,  $s$ is the scale value, and we adopt the global sum pooling operation as the relation indicator $\mathbf{M}$. 
We refer to the contrastive learning approach~\cite{zhang2021supervised}, the hyper-parameter $\tau$ mostly lies in the range of 0.08-0.1 for stable training. Therefore, the scale value $s$ in \LearningAbbr~is set to let the value learned adaptively within this range. 

In our \MainMethodAbbr, the feature pair $(x, y)$ in the loss function $L_c$ uses the output feature pairs $(\mathbf{x^a_{out}}, \mathbf{x^b_{out}})$ from \MainFusion~to calculate loss for updating the model.

For the inference, we follow Contrastive~\cite{zhang2021supervised} to
extract the final outputs $(\mathbf{x^a_{out}}, \mathbf{x^b_{out}})$ from Eq.~\ref{eq:final} to calculate the cosine similarity, and then predict whether there is a kinship relation between them with thresholding.

\begin{table*}[ht]
	\begin{center}
	\scalebox{1}{
		\begin{tabular}{c|c|c|c|c|c|c|c|c}
			\hline			
			Method &BB &SS &SIBS &FD &MD &FS &MS  &AVG.  \\
			\hline
   \hline
			\multicolumn{9}{c}{(a) Pre-trained model: ArcFace \cite{deng2019arcface}} \\ \hline

			Stefhoer$\dagger$ \cite{hormann2020multi}                         &0.660  &0.650    &0.760  &\underline{0.770}  &0.770  &0.800  &\underline{0.780} &0.740 \\
			DeepBlueAI$\dagger$ \cite{luo2020challenge}      &0.770  &0.770	&0.750	&0.740	&0.750	&0.810	&0.740	&0.760  \\   
			Ustc-nelslip$\dagger$ \cite{yu2020deep}          &0.750  &0.740	&0.720	&0.760	&0.750	&0.820	&0.750	&0.760  \\
			Vuvko$\dagger$ \cite{shadrikov2020achieving}     &0.800  &0.800	&0.770	&0.750	&0.780	&0.810	&0.740	&0.780  \\
			Contrastive \cite{zhang2021supervised}  &\underline{0.803}  &\underline{0.829}	&\underline{0.794}	&0.753	&\underline{0.803}	&\underline{0.823}	&0.751	&\underline{0.793} \\

			\textbf{\MainMethodAbbr~(Ours)}	              &\textbf{0.820}  &\textbf{0.833}	&\textbf{0.810}	&\textbf{0.773}	&\textbf{0.804}	&\textbf{0.826}	&\textbf{0.788}	&\textbf{0.806} \\ \hline\hline
			\multicolumn{9}{c}{(b) Pre-trained model: AdaFace \cite{kim2022adaface}} \\ \hline

			Contrastive \cite{zhang2021supervised} (naive) &0.630	&0.776	&0.731	&0.663	&0.687	&0.736	&0.687	&0.728 \\
			Contrastive \cite{zhang2021supervised}   &\underline{0.821}	&\underline{0.831}	&\underline{0.798}	&\underline{0.766}	&\underline{0.806}	&\underline{0.828}	&\underline{0.767}	&\underline{0.802} \\
			\textbf{\MainMethodAbbr~(Ours)}	&\textbf{0.832}	&\textbf{0.836}	&\textbf{0.824}	&\textbf{0.795}	&\textbf{0.818}	&\textbf{0.848}	&\textbf{0.802}	&\textbf{0.820} \\		\hline
		\end{tabular}}

	\end{center}
 	\caption{The state-of-the-art performance comparison of \textit{Kinship Verification} on FIW dataset by two pre-trained backbones: (a) ArcFace \cite{deng2019arcface} and (b) AdaFace \cite{kim2022adaface}. 
 The best and second results in each column are in \textbf{bold} and \underline{underline}, respectively. 
 $\dagger$The results are from \cite{robinson2021survey}.}
 	\label{tab:result1}
\end{table*}

%% file: 4_experiment.tex
\subsection{Datasets and Evaluation}
The compared methods are trained and tested on three publicly available kinship recognition datasets: \textbf{Families in the Wild (FIW)} \cite{robinson2021survey}, and \textbf{KinFaceW-I and II} \cite{lu2013neighborhood}.

The \textbf{FIW} dataset is the \textit{largest} kinship recognition dataset which includes $1000$ different and disjoint family trees, around $12000$ family photos, and $11$ kin relationship types. All face images are cropped to the size of $112 \times 112$ with face detection and alignment in training and testing by MTCNN \cite{zhang2016joint}. The 11 kin relationship types include: a) \textit{Siblings}: Brother-Brother (BB), Sister-Sister (SS), and Sister-Brother (SIBS); b) \textit{Parent-Child}: Father-Daughter (FD), Mother-Daughter (MD), Father-Son (FS), and Mother-Son (MS); c) \textit{Grandparent-Grandchild}: GFGD, GFGS, GMGD, and GMGS, with the same naming convention as above. In this work, we mainly focus on the first 7 kinship relationships since the Grandparent-grandchild categories contain much smaller data by an order of magnitude. The evaluation of FIW comprises three tasks: 1) \textit{Kinship Verification (one-to-one)}: verify the kinship relationship to predict whether a pair of individuals are blood relatives; 2) \textit{Tri-Subject Verification (one-to-two)}: the goal is to determine whether a child is related to a pair of parents; 3) \textit{Search and Retrieval (many-to-many)}: the goal is to find images in the gallery (31,787 images) that are most likely to have a kinship with the probe (190 families). For evaluation, we adopt cosine similarity and thresholding to calculate accuracy according to the FIW benchmark \cite{robinson2021survey}.

The \textbf{KinFaceW-I and II} datasets are two widely-used kinship datasets for evaluation, which include 4 kin relationship types include: Father-Son (FS), Father-Daughter (FD), Mother-Son (MS), and Mother-Daughter (MD). The KinFaceW-I dataset contains 134 (FS), 156 (FD), 127 (MS), and 116 (MD) pairs of parent-child facial images. The KinFaceW-II dataset consists of 250 pairs of facial images for each kinship relation. For evaluation, we adopt the five-fold cross-validation in the experiments following the standard protocol in GKR~\cite{li2020graph}.

\subsection{Implementation Details} \label{sec:implementation}
For experiments, we select $103784$ positive and negative image pairs overall without non-aligned images for the training phase and follow the evaluation protocols as detailed in \cite{robinson2017visual} by applying the restricted protocol where the identities of the subjects in the dataset are unknown, and we are given predefined pairs of training images, per kinship class. 
We compare our \MainMethodAbbr~against several existing methods by using ArcFace \cite{deng2019arcface} as the pre-trained backbone for a fair comparison. To demonstrate the advanced face feature representation for kinship recognition, we use the SOTA face recognition model, AdaFace \cite{kim2022adaface}, as the feature extraction network to compare with SOTA kinship recognition methods. Since the naive pre-trained weights of Adaface are not suitable for the kinship method (more details in Sec.~\ref{sec:SOTA}, we modified the initialization model parameters as a normal distribution, with lower and upper bound to [-0.05, 0.05] and utilize the L2-norm feature normalization. For the training scheme, we use SGD as the optimizer with a constant learning rate of 1e-4 and a momentum of 0.9. The batch size is set as 50, and the models are trained for 50 epochs. 
The scale value $s$ in Relation-Guided Contrastive Loss is set to 500 for stable training.

\subsection{Experimental Results}

\subsubsection{Comparison to SOTA Methods} \label{sec:SOTA}
% For objective quality assessment, 
We first evaluate kinship recognition performance on the FIW dataset for the three tasks given in the standard protocol, and compare our method with several state-of-the-art methods including stefhoer \cite{hormann2020multi}, DeepBlueAI \cite{luo2020challenge}, Vuvko \cite{shadrikov2020achieving}, Ustc-nelslip \cite{yu2020deep}, and Contrastive \cite{zhang2021supervised}. 
% In order to verify the robustness and generality of the kinship algorithm, 

Table~\ref{tab:result1} compares the \textit{Kinship Verification (one-to-one)} accuracy by using two different pre-trained models (i.e., ArcFace and AdaFace) by various methods. The result shows that the kinship recognition average accuracy from our proposed method is significantly higher than those achieved by the other methods. For the standard comparison which adopts Arcface \cite{deng2019arcface} as the pre-trained model, our \MainMethodAbbr~outperforms previous leading methods Ustc-nelslip, Vuvko, and Contrastive by 4.6 percent (0.760 $\rightarrow$ 0.806), 2.6 percent (0.780 $\rightarrow$ 0.806), and 1.3 percent  (0.793 $\rightarrow$ 0.806), respectively, as shown in Table~\ref{tab:result1}(a). 
% To demonstrate the generality of \MainMethodAbbr, 
Then a question arises: \textit{Do advanced face recognition models benefit kinship recognition?} To answer this, we adopt Contrastive \cite{zhang2021supervised} as the strong baseline and
exploit AdaFace \cite{kim2022adaface} as pre-trained for better general initial face representation. However, naively replacing the pre-trained model with Adaface is not suitable for kinship recognition, as the average accuracy decrease significantly (0.793 $\rightarrow$ 0.728). 
% This also demonstrates that contrastive learning is very sensitive to hyper-parameter with heuristic tuning.
We then modify the training scheme as stated in Sec.~\ref{sec:implementation} and show that advanced face recognition models can improve the kinship recognition task (0.793 $\rightarrow$ 0.802). Finally, by integrating the modified AdaFace backbone with our proposed \MainMethodAbbr, the result is further boosted by 1.8 percent (0.802 $\rightarrow$ 0.820), achieving the SOTA performance, as shown in Table~\ref{tab:result1}(b).
% , demonstrates that our \MainMethodAbbr~outperforms the SOTA method by 1.8 percent in average accuracy.  
% To summarize, by integrating the advanced face recognition model with \MainMethodAbbr~and our proposed training scheme, our result significantly outperforms the previous SOTA method by 2.7 percent (0.793 $\rightarrow$ 0.820), achieving a new SOTA result. 

Table~\ref{tab:track2} compares the \textit{Tri-Subject Verification (one-to-two)} performance by using the AdaFace to be the pre-trained model. The results demonstrate that the average accuracy from our \MainMethodAbbr~outperforms previous leading methods DeepBlueAI, Ustc-nelslip, and Contrastive by 8.1 percent (0.770 $\rightarrow$ 0.851), 6.1 percent (0.790 $\rightarrow$ 0.851), and 0.7 percent  (0.844 $\rightarrow$ 0.851), respectively. 

In the \textit{Search and Retrieval (many-to-many)} task, the problem is transformed into a many-to-many verification task, significantly increasing the task difficulty.
% where the gallery consists of 31,787 facial images from 190 families, and the model outputs ranked lists of all faces in the gallery that are potentially related to the input image. 
We apply our proposed \MainMethodAbbr~model from the kinship verification task (i.e., the pre-trained model from AdaFace and train from the verification task).  Table~\ref{tab:track3} demonstrates that the search and retrieval performance from our \MainMethodAbbr~significantly outperforms current SOTA methods Vuvko, Ustc-nelslip, and Contrastive by 16.2 percent (0.390 $\rightarrow$ 0.542), 31.2 percent (0.230 $\rightarrow$ 0.542), and 14.2 percent  (0.400 $\rightarrow$ 0.542), respectively. These results show that the extracted face componential relation information by our proposed \MainMethodAbbr~substantially benefits the challenging many-to-many task.

To summarize, our \MainMethodAbbr~with the proposed training scheme
% by integrating the advanced face recognition model with \MainMethodAbbr~and our proposed training scheme, our result 
significantly outperforms the SOTA methods on all three tasks in the FIW benchmark, achieving new SOTA results.

We also evaluate the kinship verification performance of our method on two widely-used databases: KinFaceW-I \cite{lu2013neighborhood} and KinFaceW-II \cite{lu2013neighborhood}, and compare our proposed \MainMethodAbbr~with several state-of-the-art methods including MNRML \cite{lu2013neighborhood}, MNRML \cite{lu2013neighborhood}, DMML \cite{yan2014discriminative}, CNN-Basic \cite{zhang122015kinship}, CNN-Point \cite{zhang122015kinship}, D-CBFD \cite{yan2019learning}, WGEML \cite{liang2018weighted}, GKR \cite{li2020graph} and Contrastive \cite{zhang2021supervised}. Table~\ref{tab:kinface} compares the kinship verification performance by using the pre-trained ResNet-18 in various methods. The result shows that the kinship verification accuracy from our proposed method is comparable to or higher than those achieved by the other methods.
% , even though the datasets KinFace-I and KinFace-II comprise only 533 and 1,000 pairs of kinship images, respectively.

\begin{table}[ht]
	\begin{center}
  	% \vspace{-0.1in}
	\scalebox{1}{
		\begin{tabular}{c|c|c|c}
			\hline			
			Method &FMD &FMS  &AVG.  \\
			\hline\hline
			Stefhoer$\dagger$ \cite{hormann2020multi}         &0.720  &0.740 &0.730 \\
			DeepBlueAI$\dagger$ \cite{luo2020challenge}      	&0.760	&0.770	&0.770  \\   
			Ustc-nelslip$\dagger$ \cite{yu2020deep}          	&0.780	&0.800	&0.790  \\
			Contrastive \cite{zhang2021supervised}          	&\underline{0.824}	&\textbf{0.860}	&\underline{0.844} \\
			\textbf{\MainMethodAbbr~(Ours)}	            &\textbf{0.841}	&\underline{0.857}	&\textbf{0.851} \\ \hline
		\end{tabular}}
	\end{center}
  	\caption{The state-of-the-art performance comparison of \textit{Tri-subject Verification} on FIW dataset. 
   The best and second results are in \textbf{bold} and \underline{underline}, respectively.
   $\dagger$The results are from \cite{robinson2021survey}.
   }
	\label{tab:track2}
\end{table}

\begin{table}[ht]
	\begin{center}

	\scalebox{1}{
		\begin{tabular}{c|c|c}
			\hline			
			Method &Rank@5 &AVG.  \\
			\hline\hline
			Vuvko$\dagger$ \cite{shadrikov2020achieving}      &\underline{0.600}	&0.390  \\   
			DeepBlueAI$\dagger$ \cite{luo2020challenge}       &0.320	&0.190  \\   
			Ustc-nelslip$\dagger$ \cite{yu2020deep}          	&0.380	&0.230  \\
			Contrastive \cite{zhang2021supervised}          	&\underline{0.600}	&\underline{0.400} \\
			\textbf{\MainMethodAbbr~(Ours)}	            &\textbf{0.668}	&\textbf{0.542} \\ \hline
		\end{tabular}}
	\end{center}
  	\caption{The state-of-the-art performance comparison of \textit{Search and Retrieval} on FIW dataset. 
   The best and second results are in \textbf{bold} and \underline{underline}, respectively. 
   $\dagger$The results are from \cite{robinson2021survey}.
   }
	\label{tab:track3}
\end{table}

\begin{table}
	\begin{center}
	\scalebox{0.9}{
		\begin{tabular}{c|c|c|c|c|c}
			\hline			
			Method &FS &FD &MS &MD  &AVG. \\
			\hline\hline
			\multicolumn{6}{c}{KinFaceW-I} \\ \hline
			MNRML$\dagger$ \cite{lu2013neighborhood}           &0.725 &0.665 &0.662 &0.720 &0.699 \\
	        DMML$\dagger$  \cite{yan2014discriminative}        &0.745 &0.695 &0.695 &0.755 &0.723 \\
	        CNN-Basic$\dagger$ \cite{zhang122015kinship}       &0.757 &0.708 &0.734 &0.794 &0.748 \\
	        CNN-Point$\dagger$ \cite{zhang122015kinship}       &0.761 &0.718 &0.780 &\underline{0.841} &0.775 \\
	        D-CBFD$\dagger$ \cite{yan2019learning}             &0.790 &0.742 &0.754 &0.773 &0.785 \\
	        WGEML$\dagger$  \cite{liang2018weighted}           &0.785 &0.739 &0.806 &0.819 &0.787 \\
	        GKR$\dagger$ \cite{li2020graph}                    & \underline{0.795} &0.732 &0.780 &\textbf{0.862} &0.792 \\
	        Contrastive \cite{zhang2021supervised}  &\textbf{0.799}  &\underline{0.805}  &\underline{0.835}  & 0.780 &\underline{0.805} \\
		    \textbf{\MainMethodAbbr~(Ours)}    &\textbf{0.799} &\textbf{0.818} &\textbf{0.839} &0.806 &\textbf{0.815} \\
            \hline\hline
			\multicolumn{6}{c}{KinFaceW-II} \\ \hline
			MNRM$\dagger$L \cite{lu2013neighborhood}           &0.769 &0.743 &0.774 &0.776 &0.765 \\
	        DMML$\dagger$ \cite{yan2014discriminative}         &0.785 &0.765 &0.785 &0.795 &0.783 \\
	        CNN-Basic$\dagger$ \cite{zhang122015kinship}       &0.849 &0.796 &0.883 &0.885 &0.853 \\
	        CNN-Point$\dagger$ \cite{zhang122015kinship}       &\underline{0.894} &0.819 &0.899 &\underline{0.924} &0.884 \\
	        D-CBFD$\dagger$ \cite{yan2019learning}             &0.810 &0.762 &0.774 &0.793 &0.785 \\
	        WGEML$\dagger$ \cite{liang2018weighted}            &0.886 &0.774 &0.834 &0.816 &0.828 \\
	        GKR$\dagger$   \cite{li2020graph}                  &\textbf{0.908} &0.860 &\textbf{0.912} &\textbf{0.944} &\textbf{0.906} \\
	        Contrastive \cite{zhang2021supervised}  &0.852  &\underline{0.898}  &\textbf{0.912}  & 0.890 &\underline{0.888} \\
		    \textbf{\MainMethodAbbr~(Ours)}    &0.886 &\textbf{0.922} &\underline{0.910} &0.900 &\textbf{0.906} \\
            \hline
		\end{tabular}}

	\end{center}
  	\caption{Verification accuracy of different methods on KinFaceW-I and KinFaceW-II datasets. 
      The best and second results are in \textbf{bold} and \underline{underline}, respectively.
   $\dagger$The results are from \cite{li2020graph}.}
	\label{tab:kinface}
\end{table}

\begin{table*}
	\begin{center}
	\scalebox{1}{
		\begin{tabular}{c|c|c|c|c|c|c|c|c}
			\hline			
			Method &BB &SS &SIBS &FD &MD &FS &MS &AVG.  \\
			\hline\hline
			\multicolumn{9}{c|}{(a) Standard Protocol} \\ \hline
			Contrastive \cite{zhang2021supervised}  &0.803	&0.829	&0.794	&0.753	&0.803	&0.823	&0.751	&0.793 \\
			\textbf{\MainMethodAbbr~(Ours)}	&\textbf{0.832}	&\textbf{0.836}	&\textbf{0.824}	&\textbf{0.795}	&\textbf{0.818}	&\textbf{0.848}	&\textbf{0.802}	&\textbf{0.820} \\ \hline\hline
			\multicolumn{9}{c|}{(b) Quality-Filtered Protocol (Quality Score $>$ 0.5)} \\ \hline
			Contrastive \cite{zhang2021supervised}  &0.800	&0.817	&0.772	&0.739	&0.784	&0.836	&0.786 &0.792 \\
			\textbf{\MainMethodAbbr~(Ours)}	&\textbf{0.836}	&\textbf{0.838}	&\textbf{0.784}	&\textbf{0.784}	&\textbf{0.842}	&\textbf{0.862}	&\textbf{0.815}	 &\textbf{0.826}\\
			\hline
		\end{tabular}}

	\end{center}
   	\caption{Performance comparison of kinship on FIW dataset by using AdaFace to be the pre-trained model in two quality-filtered protocols: (a) standard protocol: use all image pairs without filtering; (b) quality-filtered protocol: select the image pairs with the pair quality scores larger than 0.5, which is more practical in real-world scenarios.
   }
	\label{tab:result2}
\end{table*}

\subsubsection{Practical Kinship Recognition Protocol}
With the improvement of hardware, the photos captured by cameras or smartphones have better quality, so it is worth investigating the impact of using higher-quality face images in practical applications. We conduct an experiment for practical kinship recognition as shown in Table~\ref{tab:result2} (b). More specifically, we propose a quality-filtered protocol, where we select high-quality training and testing face images with SER-FIQ quality scores \cite{terhorst2020ser} larger than 0.5. The results demonstrate that the average accuracy of \MainMethodAbbr~are significantly higher than the baseline (i.e., Contrastive). This trend is similar to the standard protocol as shown in Table~\ref{tab:result2} (a), but the improvement from our method over the baseline is even more obvious (0.792 $\rightarrow$ 0.826).

Intuitively, using high-quality face images as training and testing data would improve overall accuracy. However, the accuracy of Contrastive~\cite{zhang2021supervised} does not improve on high-quality face images, which also confirms that our \MainMethodAbbr~can learn the correlation between image pairs and fuse them more effectively, that is, capture the face components from the eye, nose, and mouth. Besides, we further analyze the recognition results of different kinships and found that the accuracy of the same gender (i.e., BB, SS, MD, FS) was significantly higher. Among them, the result of \MainMethodAbbr~+ \LearningAbbr~in MD case has a significant improvement of 2.4 percent (0.818 $\rightarrow$ 0.842) from the standard to the quality-filtered protocol, showing that the MD cases include a large amount of low-quality face images in the standard protocol. 
On the other hand, MD has slightly lower recognition accuracy than FS, and we conjecture that it is due to the challenging MD cases caused by makeup and coverings. 
Moreover, the accuracy of the SIBS case decreases after selecting high-quality face images. The main reason is that SIBS has less data than other kinship categories. Finally, the results also demonstrate that our \MainMethodAbbr~outperforms the SOTA method by a large margin in all kin categories.

\subsubsection{Ablation Studies}
\noindent\textbf{Component Analysis:}
In this section, we conduct an ablation study to analyze the proposed design for comparisons against various component modules in \textit{Kinship Verification}, \textit{Tri-subject Verification}, and \textit{Search and Retrieval} tasks on the FIW dataset, corresponding to task 1, task 2, and task 3, respectively. Our proposed \MainMethodAbbr~utilizes the \MainFusion~(\MainFusionAbbr) to extract the face components of both images in a pair by assigning the important facial regions for kinship recognition. Table~\ref{tab:result3} reports the contributions of individual modules of \MainMethodAbbr. The first column shows the naive approach of contrastive learning with only 2 fully-connected layers from \cite{zhang2021supervised}. The second column shows the improvement of utilizing \cite{zhang2021supervised} with our proposed \MainFusionAbbr~module. The result demonstrates that our \MainFusionAbbr~module significantly improves, which reveals that \MainFusionAbbr~can extract the face component information by learning the correlation between image pairs. 

The last column shows that the relation guidance in contrastive loss can improve kinship performance. This was achieved by using \textit{\LearningLoss~(\LearningAbbr)} with a relation indicator, which automatically estimate $\tau$ value in a certain range in the contrastive loss instead of heuristic tuning of fixed $\tau$, particularly on the challenging task 3.

\begin{table}[ht]
	\begin{center}
	\scalebox{0.9}{
		\begin{tabular}{c|c|c||c|c|c}
			\hline		
                Contrastive & \MainFusionAbbr &  \LearningAbbr & task 1& task 2 &task 3  \\ \hline\hline
			\checkmark &  & 	&0.793 & 0.844 & 0.400\\

			\checkmark & \checkmark & 	&0.803 & 0.848 & 0.511 \\
			\checkmark & \checkmark & \checkmark	&\textbf{0.806} &\textbf{0.851} &\textbf{0.542} \\ \hline
		\end{tabular}}

	\end{center}
  	\caption{Component analysis of \MainMethodAbbr on the FIW dataset, including \textit{Kinship Verification}, \textit{Tri-subject Verification}, and \textit{Search and Retrieval} tasks, corresponding to task 1, task 2, and task 3, respectively.}
	\label{tab:result3}
\end{table}

\noindent\textbf{Face Quality Analysis:}
We perform an experiment to evaluate accuracy of applying different methods to face images with different image qualities. We use SER-FIQ \cite{terhorst2020ser} to compute the face quality scores of all images and adopt the lower score in a pair as the face-pair quality score. We divide the face-pair quality scores into 5 groups as shown in Table~\ref{tab:result4}. The results show that in low-quality cases (i.e., the quality scores$<$0.4), the overall recognition accuracy is lower than those in high-quality cases. The problem is more severe in extremely low-quality cases (i.e., 0.2). 
Finally, the results also demonstrate that our \MainMethodAbbr~ outperforms the SOTA method under all different levels of face qualities.

\begin{table}[ht]
	\begin{center}
	\scalebox{0.85}{
		\begin{tabular}{c|c|c}
		\hline
        Face-Pair Quality Score &Contrastive \cite{zhang2021supervised} &\textbf{\MainMethodAbbr~(Ours)} \\ \hline\hline
        0-0.2    &0.749 &\textbf{0.794}\\
        0.2-0.4  &0.782 &\textbf{0.820}\\
        0.4-0.6  &0.813 &\textbf{0.843}\\
        0.6-0.8  &0.803 &\textbf{0.821}\\
        0.8-1   &0.793 &\textbf{0.824}\\ \hline
        AVG.    &0.793 &\textbf{0.820}\\ \hline
		\end{tabular}}
	\end{center}
  	\caption{Performance comparison of kinship on FIW dataset under various groups of pair quality scores. The table represents the pair quality score in groups from small to large.}
	\label{tab:result4}
\end{table}

\noindent\textbf{Face Verification:}
We also evaluate face verification performance with two databases: LFW \cite{huang2008labeled} and AgeDB-30 \cite{moschoglou2017agedb}, and compare our proposed \MainMethodAbbr~with two methods including the state-of-the-art Contrastive \cite{zhang2021supervised} approach and the regular face recognition model (Arcface-MS1M \cite{deng2019arcface}). The regular face recognition model of Arcface-MS1M is trained by using the MS1MV3 \cite{pernici2019maximally} database, which represents the upper bound of face recognition. As shown in Table~\ref{tab:face_verf}, the results show that the face verification performance of our proposed method is significantly higher than the Contrastive scheme. It demonstrates that our proposed method not only learns the discriminative feature of kinship recognition but also retains the identity information.

\begin{table}
	\begin{center}

	\scalebox{1}{
		\begin{tabular}{c|c|c}
			\hline
			Method &ACC. &AUC  \\ \hline\hline
			\multicolumn{3}{c}{(a) LFW dataset} \\ \hline
			ArcFace-MS1M$\dagger$ \cite{deng2019arcface}                    &0.998	&0.999 \\ \hline
			Contrastive \cite{zhang2021supervised}          	&0.993	&0.998 \\
			\textbf{\MainMethodAbbr~(Ours)}	            &\textbf{0.995}	&0.999 \\ \hline\hline
			\multicolumn{3}{c}{(b) AgeDB-30 dataset} \\ \hline
			ArcFace-MS1M$\dagger$ \cite{deng2019arcface}                    &0.980	&0.991 \\ \hline
			Contrastive \cite{zhang2021supervised}          	&0.965	&0.989 \\
			\textbf{\MainMethodAbbr~(Ours)}	            &\textbf{0.970}	&0.989 \\ \hline

		\end{tabular}}
	\end{center}
  	\caption{The results of \textit{Face Verification} comparison on LFW and AgeDB-30 dataset. $\dagger$The upper-bound of the face verification.}
	\label{tab:face_verf}
	\vspace{-0.2in}
\end{table}

\noindent\textbf{Visualization:}

\begin{figure}[!hbt]
\centering
\includegraphics[width=0.5\textwidth]{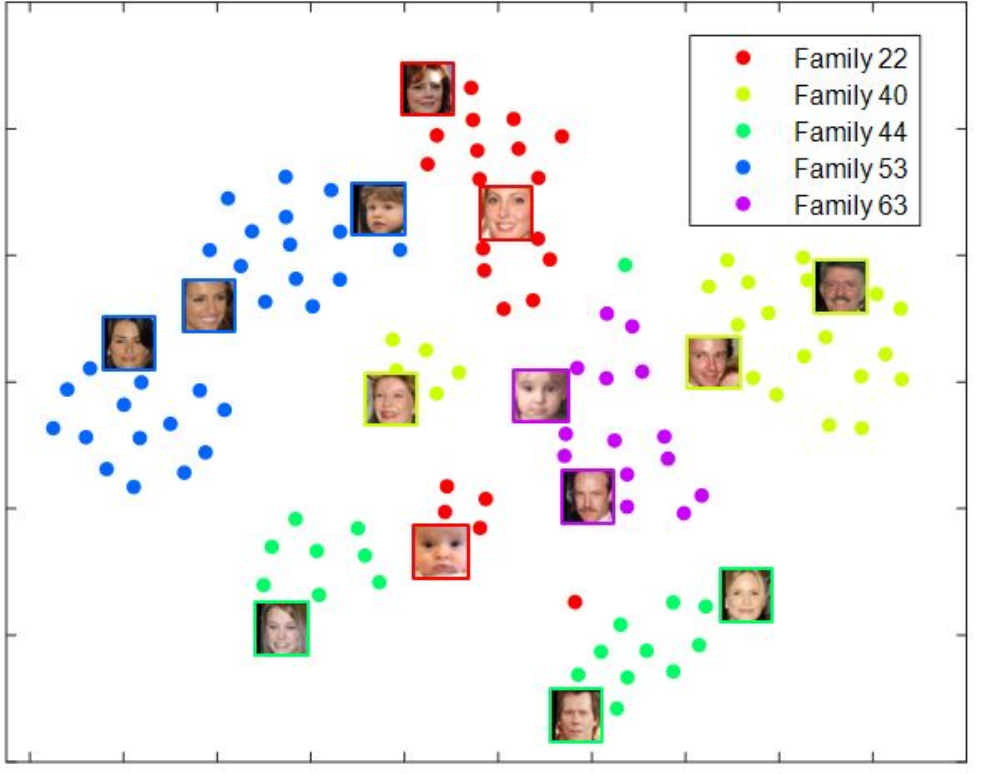}
\caption{Visual analysis of the learned feature with t-SNE.}
\label{fig:tsne}	
\vspace{-0.1in}
\end{figure}

We also perform visual analysis on the latent features learned by our proposed model, as shown in Fig.~\ref{fig:tsne}. We select 5 families from the FIW validation set and visualize the distribution by using t-SNE. We observe that members of the same family are close to each other, while there are gaps between members of different families. It indicates the discriminative ability of our kinship model. 

\begin{figure}[!hbt]
\centering
\includegraphics[width=0.5\textwidth]{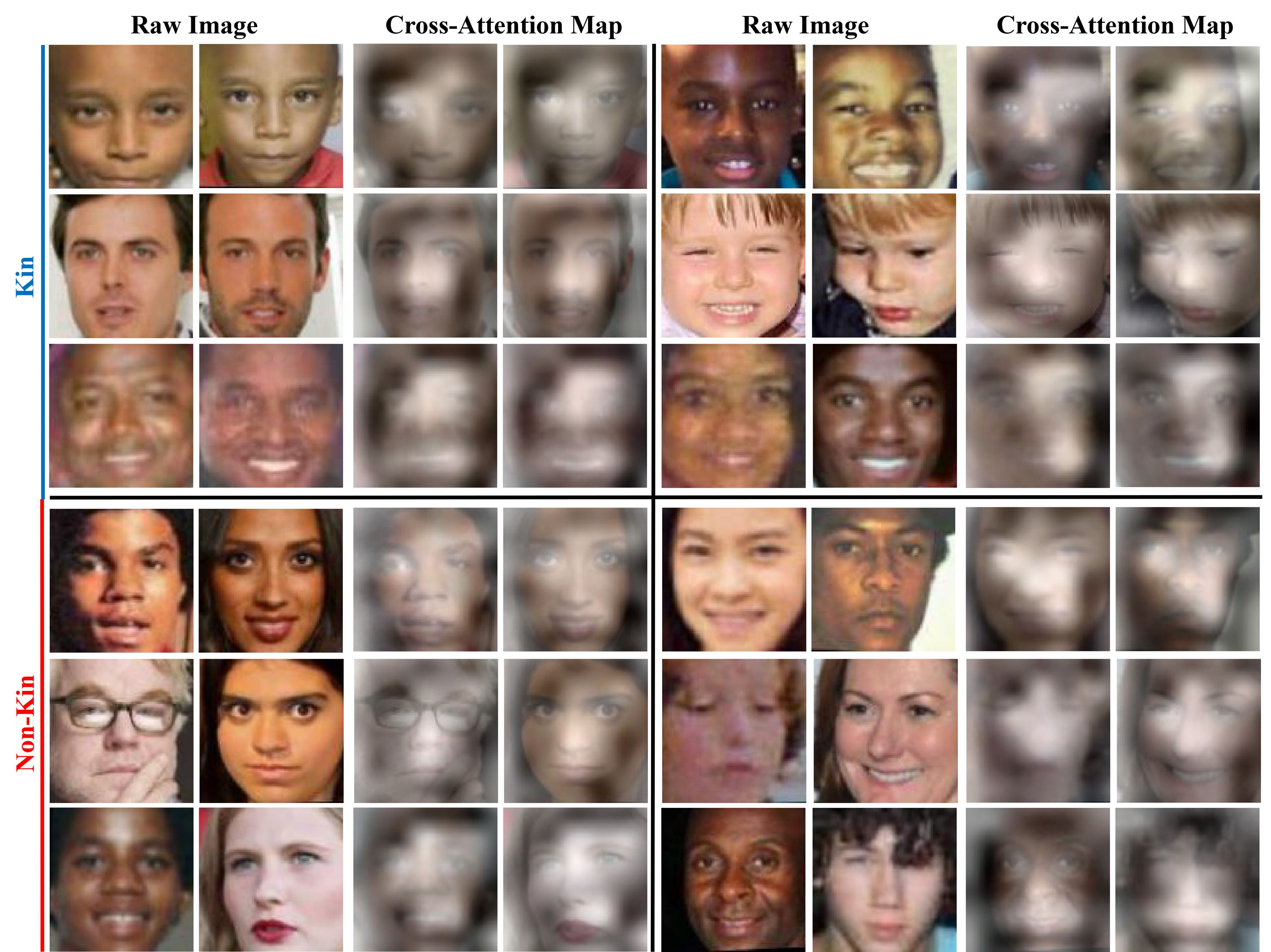}
\caption{Illustration of Cross-Attention map includes kin (first to third rows) and non-kin (fourth to sixth rows) cases: the left side shows the raw image pairs and the right side shows the visualization of the cross-attention maps.}
\label{fig:att}	
\end{figure}

Besides, Fig.~\ref{fig:att} visualizes 
the cross-attention map $\boldsymbol{\beta}$ produced by our proposed model from Eq.~\ref{eq:correlation}. 
The visualization results demonstrate that our proposed method \MainMethodAbbr~attains higher values in the face components, particularly in the regions of the eye, nose, and mouth where there is a high degree of similarity, and where the covering is skipped. These findings substantiate that our \MainMethodAbbr~method can effectively learn the relevance of the face components in image pairs through cross-attention estimation, and can adaptively acquire critical face components for accurate kinship recognition.

%% file: 5_conclusion.tex
In this paper, we propose a novel \textit{\textbf{Fa}ce \textbf{Co}mponential \textbf{R}elation \textbf{Net}work (\textbf{\MainMethodAbbr})} for kinship recognition. \MainMethodAbbr~is an attention-based model designed for learning correlation between image pairs in terms of face components. To better address large variations in facial appearance, \MainMethodAbbr~utilizes the \textit{\MainFusion~(\MainFusionAbbr)} module to achieve not only adaptive learning correlation between image pairs but also learning important face components for kinship recognition. In addition, we embed the cross-attention estimation as a relation indicator to guide the regular contrastive loss without the need for heuristic tuning.
Experimental results show that our method achieves SOTA performance on multiple kinship recognition benchmarks, including the FIW benchmark. Moreover, for practical kinship recognition protocol, \MainMethodAbbr~also outperforms the SOTA methods by large margins. We believe that \MainMethodAbbr~can be served as a strong baseline for further advancing facial relation learning approaches in kinship recognition. For future work, we plan to incorporate face quality scores into the training process, aiming to mitigate the issues from low-quality face images. We would also like to incorporate multi-modal information (e.g., text, metadata) to compensate for the vision-based methods.